\documentclass[11pt]{article}

\usepackage[margin=1in]{geometry}
\usepackage{hyperref}
\usepackage{graphicx}
\usepackage{booktabs}
\usepackage{listings}
\usepackage{xcolor}
\usepackage{amsmath}
\usepackage{parskip}
\usepackage{pifont}
\usepackage[section]{placeins}

\hypersetup{
  colorlinks=true,
  linkcolor=blue,
  citecolor=blue,
  urlcolor=blue
}

\newcommand{\yes}{\ding{51}}
\newcommand{\no}{\ding{55}}
\newcommand{\partmark}{$\sim$}

\title{\textbf{WebSerial Vision Training for Microcontrollers:\\
A Browser-Based Companion to On-Device CNN Training}\\[0.4em]
{\large webmcu-vision-web (Paper~2 of the webmcu-ai Series)}\\[4pt]
{\small \href{https://github.com/webmcu-ai/webmcu-vision-web/releases/tag/v1.0.0}{Version 1.0.0}}}

\author{
  Jeremy Ellis\\
  High School Robotics Educator\\
  British Columbia, Canada\\
  \href{https://github.com/webmcu-ai}{\texttt{https://github.com/webmcu-ai}}
}

\date{April 2026}

\begin{document}

\maketitle

\begin{abstract}
This paper presents \texttt{webmcu-vision-web}, a single-file, zero-install browser
application for end-to-end TinyML vision model training and deployment on the
Seeed Studio XIAO ESP32-S3 Sense (XIAO ML Kit, \$15--40\,USD). Acting as a browser-based companion to the
on-device Arduino firmware of Paper~1~\cite{ellis2026ondevice}, it provides a
private, fully local machine learning pipeline---from firmware flashing through image
collection, CNN training, weight export, and live activation visualization---without
any software installation beyond a Chromium-based browser. The system targets
educators, small businesses, and researchers who need to train task-specific visual
classifiers under their exact deployment conditions rather than adapting large
general-purpose datasets.

Key capabilities include: in-browser firmware flashing via \texttt{esptool-js};
a bidirectional serial monitor; a full SD card file browser with image preview and
inline editing; \texttt{config.json} live-sync for zero-recompile hyperparameter
adjustment; webcam and ESP32 OV2640 camera image capture; TensorFlow.js CNN training
matching the on-device architecture, completing a representative three-class run
($\sim$30 images per class, 20 epochs) in approximately 1~minute browser-side versus
9~minutes on-device, enabling a complete collect--train--deploy cycle in under
10~minutes; weight export as \texttt{myWeights.bin} and \texttt{myWeights.h} for
deployment to the SD card; confusion matrix; and a live Conv2 activation heatmap
streamed from the ESP32 during inference. No data leaves the local machine at any
stage. A five-run consistency evaluation on the three-class reference problem
(0Blank, 1Cup, 2Pen) with independently collected datasets demonstrates stable
convergence and provides mean accuracy with standard deviation; all artefacts and
raw results are released at the repository link below. The repository is structured as a living template designed for
LLM-assisted adaptation to new hardware, sensors, and tasks. All source code is
released under the MIT License at
\href{https://github.com/webmcu-ai/webmcu-vision-web}{\texttt{https://github.com/webmcu-ai/webmcu-vision-web}}.
\end{abstract}


\tableofcontents

\newpage

\section{Introduction}

Paper~1 of this series~\cite{ellis2026ondevice} demonstrated that a complete CNN
training pipeline---data capture, backpropagation, Adam optimization~\cite{kingma2015adam}, and 6.3~FPS
inference---executes entirely on a \$15--40 USD microcontroller with no external
computation, completing one training run ($\sim$90 images, 20 epochs) in approximately
9 minutes. That paper defines Tier~1 of a three-tier mental model for embedded ML
education (Table~\ref{tab:tiers}).

\begin{table}[h]
\centering
\caption{Three-tier ML deployment model (from Paper~1~\cite{ellis2026ondevice}).
Tier~2 occupies the pedagogical sweet spot: it retains the local-data privacy guarantee
of Tier~1 while providing the UI richness and training speed of a laptop.}
\label{tab:tiers}
\begin{tabular}{@{}lp{3.4cm}p{4.2cm}l@{}}
\toprule
\textbf{Tier} & \textbf{Infrastructure} & \textbf{When to choose} & \textbf{Privacy} \\
\midrule
1. On-device      & Microcontroller only     & Sensitive data, no connectivity,
                                              ultra-low power, transparent education
                                            & Raw data stays on-device \\[4pt]
2. Local computer & Laptop + USB (WebSerial) & Larger datasets, classroom use,
                                              faster iteration, richer UI
                                            & Data stays on local machine \\[4pt]
3. Cloud          & Remote GPU cluster       & Production scale, large teams,
                                              foundation models
                                            & Data leaves premises \\
\bottomrule
\end{tabular}
\end{table}

\subsection{The TinyML Dataset Philosophy}

A key distinction of TinyML relative to large-scale computer vision is the role of
the dataset. In cloud-based ML, model generality is a virtue: large datasets covering
diverse conditions produce models that work everywhere. TinyML inverts this priority.
A device consuming milliwatts must solve \emph{one specific task}---detecting a
particular part on a factory line under known fluorescent lighting, or identifying
a specific plant disease in a single greenhouse. The correct dataset for this task
is not a large general corpus: it is a small, clean, purposefully collected set of
images taken \emph{under exactly the conditions of deployment}.

This paper's system is designed around that philosophy. The camera is positioned
in front of the actual objects. Images are captured under the actual lighting. The
classes are named for the actual problem. The model is trained until it converges
on that specific problem, not on a benchmark. This approach yields smaller datasets
(15--100 images per class is typically sufficient), faster training, lower energy
consumption at inference time, and higher accuracy on the target task than a
general model ever could.

This matters especially for the global south and for small businesses, where access
to cloud GPU infrastructure may be intermittent or costly, and where the specific
deployment conditions differ substantially from any publicly available dataset. A
worker in a materials-sorting facility, a researcher studying local crop diseases,
a teacher in a school with unreliable internet: all can collect their own images,
train on their own hardware, and deploy a working classifier without sending any
data off-premises.

\subsection{Problem Statement}

Paper~1 of this series~\cite{ellis2026ondevice} establishes a fully self-contained
on-device ML pipeline, but operating it entirely through the device's own interface
presents practical friction points for classroom and rapid-iteration use. Firmware
installation for a new device requires the Arduino IDE with the ESP32 board package
installed. Reading collected images or adjusting training parameters means
ejecting the SD card and editing files on a PC. Any hyperparameter change requires a
source-code edit and a full firmware recompile. Capturing
training images is done through an OLED-based touch menu on the device itself, with
a very small black-and-white OLED image preview. Collectively, these friction points can consume
the majority of a class session before any ML concept is taught.

This paper addresses each of these Paper~1 workflow limitations directly with a
browser-native companion interface requiring no software installation beyond Chrome
or Edge.

\subsection{LLM-Assisted Adaptation}

This repository is explicitly designed as a template for LLM-assisted modification.
The single-file structure, consistent naming conventions, and inline documentation
are all chosen to make the codebase readable and modifiable by a practitioner using
Claude, ChatGPT, Gemini, or any other LLM assistant. A user in any country can
describe their specific hardware variant, sensor modality, or class problem to an LLM
and receive targeted diffs to the relevant sections of the file. This paper documents
the architecture precisely enough that those modifications are reproducible and
verifiable. The webmcu-ai GitHub organization is structured as a series of such
templates, each covering a different sensor modality (vision, audio, IMU), with the
goal that any researcher or small-business practitioner can adapt them with minimal
barrier.

\subsection{Contributions}

This paper makes four primary contributions:

\begin{enumerate}
  \item \textbf{A fully local, zero-install ML lifecycle for embedded systems.}
        The complete TinyML workflow---firmware flash, image collection, CNN training,
        weight transfer, and activation visualization---executes from a single HTML file
        with no installed toolchain, no cloud dependency, and no data leaving the local
        machine. This is, to our knowledge, the first such end-to-end system for
        microcontroller-class hardware delivered entirely in a browser.

  \item \textbf{A new pedagogical model for TinyML iteration.} Separating the fast
        browser loop ($\sim$1~min per run, WebGL-accelerated) from the slow embedded loop
        ($\sim$9~min per run, ESP32 FPU) creates a two-speed development model: practitioners
        iterate on dataset and hyperparameters at browser speed, then verify and deploy at
        device speed. A complete collect--train--deploy cycle takes under 10~minutes,
        supporting multiple full iterations within a single class period.

  \item \textbf{A deployment philosophy for intentional, task-specific overfitting.}
        TinyML classifiers are not intended to generalise beyond their deployment environment;
        they are designed to overfit to a specific object, under specific lighting, at a specific
        position. This is a feature, not a limitation. The data collection workflow, camera
        positioning, and evaluation methodology in this system are built around this principle,
        making it directly applicable to global south and small-business contexts where
        cloud-scale datasets do not match local conditions.

  \item \textbf{Live on-device activation heatmap streaming.}
        A Conv2 heatmap protocol streams spatial activation maps from the running ESP32
        to the browser in real-time, making the network's attention visible to the
        practitioner during inference. To our knowledge, this is one of the first
        real-time activation visualizations deployed on a microcontroller-class device.
        A complementary browser-side heatmap computed from the TensorFlow.js model
        is also available without device connection.
\end{enumerate}

Supporting capabilities enabling these contributions include: SD card reading over serial; \texttt{config.json} live-sync for zero-recompile hyperparameter
adjustment; exact weight format transpositions between TensorFlow.js and ESP32 C++ loop
order; and a single-file repository structure designed for LLM-assisted adaptation to
new hardware and sensor modalities.

\section{Related Work}

\textbf{Edge Impulse}~\cite{edgeimpulse} provides a professional cloud-based pipeline
for TinyML: data management, training, and model export to a wide range of embedded
targets. It requires account creation, uploads data to a remote server, and does not
support on-device training or fully local operation. It is the correct choice for
production-grade embedded ML at scale; the present work targets the complementary use
case of zero-install, fully local operation where data privacy and offline access are
priorities.

\textbf{Google Teachable Machine}~\cite{teachablemachine} enables browser-based image
classification training using TensorFlow.js with a minimal drag-and-drop interface.
It does not support WebSerial communication, embedded firmware deployment, SD card
management, or on-device inference. Its target audience is non-technical users rather
than engineering students or practitioners exploring the full ML system stack.

\textbf{Arduino IDE~\cite{arduinoide} and TensorFlow Lite Micro~\cite{tflitemicro}.} The standard embedded ML workflow for microcontrollers typically involves training a model in Python, converting it to a C array using TensorFlow Lite, and embedding that array in a C++ firmware sketch for deployment. This is a well-supported path and microcontrollers can also run Python via MicroPython or CircuitPython, though these sacrifice the performance of native C++. ML inference can additionally be written directly in C++ without a framework. The fragmentation arises from the toolchain: every model iteration requires a separate conversion step and a full firmware recompile. The system described here reduces iteration friction by allowing browser-side training and weight export in a format the firmware loads directly at boot, without any Python conversion step.

\textbf{WebSerial and WebUSB educational platforms.} The Web Serial
API~\cite{w3cserial} has been used for browser-based microcontroller communication in
tools such as Arduino Web Editor and MicroPython's WebREPL, but these focus on code
editing and serial monitoring rather than ML training or embedded weight deployment.

\textbf{TinyML4D}~\cite{plancher2024tinymld} scales embedded ML education globally
through open curricula, emphasising the importance of the full lifecycle. The system
described here provides a concrete, low-cost, zero-install artefact suitable for
TinyML4D-style curricula where software installation is restricted or unreliable.

Table~\ref{tab:platform} summarises the qualitative comparison. The comparison emphasises workflow characteristics rather than absolute capability; tools like Edge Impulse provide significantly broader production support and should be the default choice for teams with cloud access and production-scale deployment requirements.

\begin{table}[h]
\centering
\caption{Qualitative comparison of embedded ML development approaches.}
\label{tab:platform}
\begin{tabular}{@{}lccccc@{}}
\toprule
\textbf{Feature} & \textbf{This work} & \textbf{Edge Impulse} & \textbf{Teachable Machine} & \textbf{Arduino + TFLite} \\
\midrule
Zero software install    & \yes & \no  & \yes & \no  \\
Fully local data         & \yes & \no  & \no  & \yes \\
Browser-side training    & \yes & \no  & \yes & \no  \\
On-device training       & \yes & \no  & \no  & \no  \\
SD card reading          & \yes & \no  & \no  & \no  \\
Firmware flash (.ino / browser) & \yes & \yes  & \no  & \yes  \\
Activation heatmap       & \yes & \no  & \no  & \no  \\
Config without recompile & \yes & \yes & \yes & \no  \\
LLM-adaptable template   & \yes & \no  & \no  & \partmark \\
Production-grade targets & \no  & \yes & \partmark & \yes \\
\bottomrule
\end{tabular}
\end{table}

\section{System Overview}

\subsection{Relationship to Paper~1}

\begin{figure}[htbp]
  \centering
  \includegraphics[width=0.65\textwidth]{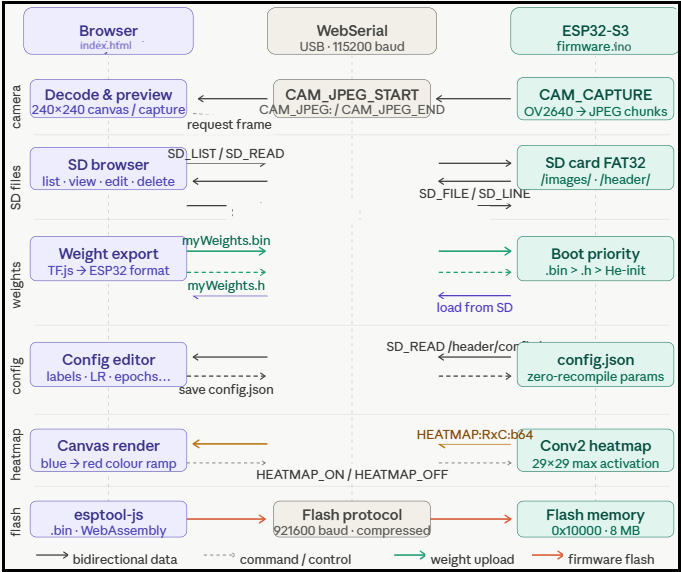}
  \caption{Data flow between \texttt{webmcu-vision-web} and the ESP32-S3 firmware.
           All communication passes over a single USB serial connection at 115200~baud
           via the browser WebSerial API. Six channels are multiplexed over this
           connection: camera frames (inbound), SD file operations (bidirectional),
           weight and config files (outbound), heatmap activation frames (inbound),
           and firmware flash (outbound). No data leaves the local machine.}
  \label{fig:overview}
\end{figure}

The on-device firmware (Paper~1) is the authoritative training engine for Tier~1
deployments. \texttt{webmcu-vision-web} is not a replacement: it is a complementary
interface that enables the same workflow through a browser when a laptop with a USB
connection is available. The two approaches share the same weight format
(\texttt{myWeights.bin}, \texttt{myWeights.h}),
and the same SD card directory layout, so weights trained in the browser are directly
deployable on the device and vice versa. Figure~\ref{fig:overview} shows the overall
data flow.

\subsection{Architecture and Single-File Design}

The application is a single HTML file (\texttt{index.html}) of approximately 4,000
lines. No framework, build step, or package manager is required. External dependencies
are loaded at runtime from public CDNs:

\begin{itemize}
  \item \textbf{TensorFlow.js 4.22.0}~\cite{tensorflowjs} --- browser-side CNN training
        and inference, GPU-accelerated via WebGL, matching the two-layer on-device
        architecture.
  \item \textbf{JSZip 3.10.1} --- client-side ZIP creation for bulk image export.
  \item \textbf{esptool-js 0.5.7}~\cite{esptooljsrepo} --- WebAssembly port of
        \texttt{esptool.py} for in-browser firmware flashing.
\end{itemize}

All application logic---WebSerial communication, SD browser, training loop, weight
import/export, heatmap rendering---is implemented in vanilla JavaScript within the
single file. The single-file constraint is deliberate: it maximises accessibility,
eliminates dependency management as a barrier, and makes the complete source
directly auditable in a browser's developer tools. It also makes LLM-assisted
modification practical: the entire codebase fits in a single context window, and
targeted changes to any section can be described precisely by reference to the
section numbering and function names documented here.

\subsection{Hardware and Software Requirements}

\begin{figure}[htbp]
  \centering
  \includegraphics[width=0.40\textwidth]{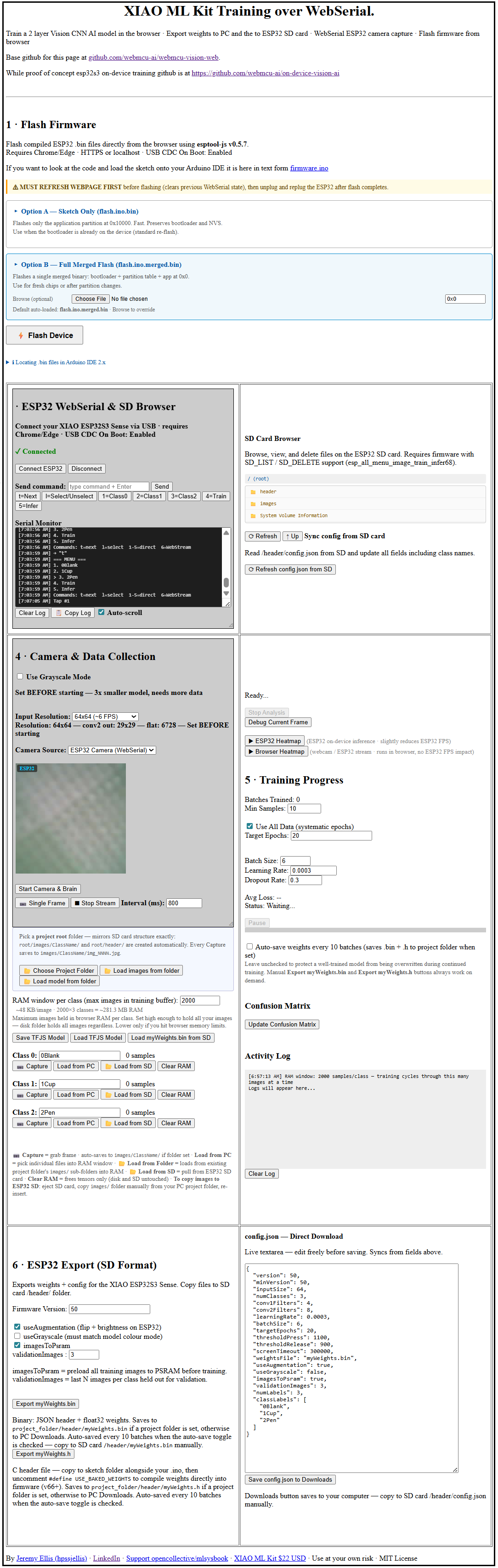}
  \caption{The complete \texttt{webmcu-vision-web} interface (\texttt{index.html}).
           All sections are visible on a single scrolling page with no navigation
           or routing. The six numbered sections (Flash Firmware, WebSerial \& SD
           Browser, Camera \& Data Collection, Training Progress, ESP32 Weight Export,
           and Activity Log) are self-contained and independently usable. The entire
           source is auditable in the browser developer tools.}
  \label{fig:fullpage}
\end{figure}

All experiments were conducted using a Seeed Studio XIAO ESP32-S3 Sense board as the XIAO ML kit
(8~MB PSRAM, OV2640 camera, MicroSD slot) running firmware~v1.0.0 of
\texttt{webmcu-vision-web}. The browser application was tested on Chrome~124 and
Edge~124 (Windows~11 and macOS~14). Figure~\ref{fig:fullpage} shows the complete
interface.

\begin{figure}[htbp]
  \centering
  \includegraphics[width=0.55\textwidth]{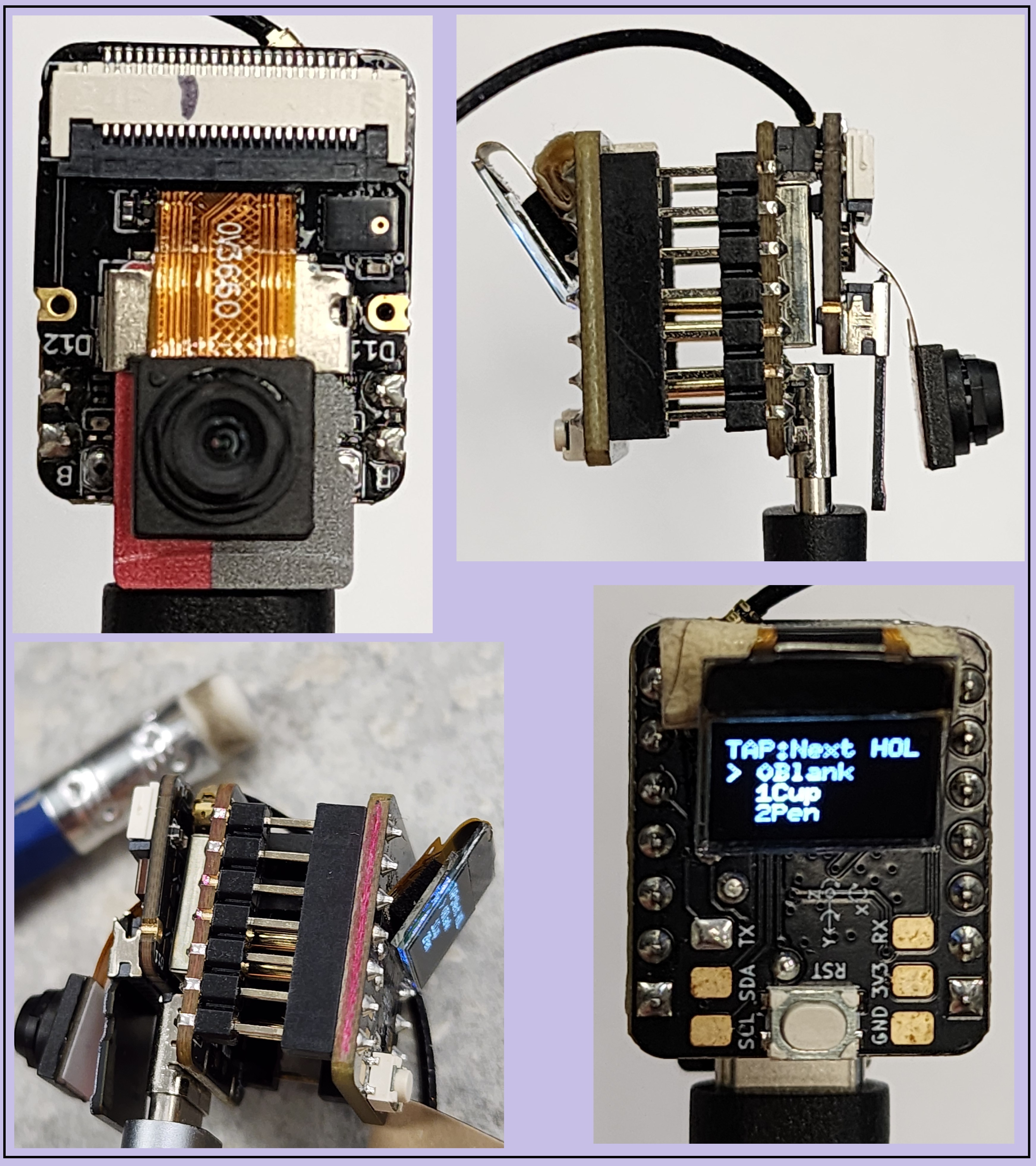}
  \caption{The XIAO ML Kit showing the ESP32-S3 Sense board with integrated OV2640
           camera, 72$\times$40 OLED display, and IMU sensor (\$38.90~USD complete
           kit; \$14.99 bare Sense board; \$8 ESP32-S3 chip alone).
           The 8~MB PSRAM is sufficient for the full browser-assisted training pipeline.
           Pin~A0 capacitive touch enables on-device menu navigation without the
           browser connection. All three hardware configurations are supported by
           the firmware; only the OLED and IMU features require the expansion board.}
  \label{fig:hardware}
\end{figure}

The target hardware is the Seeed Studio XIAO ESP32-S3 Sense (Figure~\ref{fig:hardware}).
Three entry points are available at different price levels: the bare ESP32-S3 chip
alone is available for approximately \$8~USD; the XIAO ESP32-S3 Sense board with
integrated OV2640 camera retails for approximately \$15~USD; and the full XIAO ML Kit
adding an OLED display, IMU, and expansion board is approximately \$38.90~USD.
All three configurations run the same firmware; the OLED and IMU features degrade
gracefully if the expansion board is absent. For educators and students, this price
range places a complete on-device ML platform within reach of a single classroom
budget line.

\section{Application Sections}

\subsection{In-Browser Firmware Flashing}
\label{sec:flash}

Section~1 allows educators and practitioners to flash pre-compiled ESP32 firmware
binaries directly from the browser, with no Python, \texttt{pip},
\texttt{esptool.py}, or Arduino IDE required. This eliminates the most common barrier
to first-time deployment of embedded ML systems, and is particularly significant in
environments where software installation is restricted by IT policy or unreliable
internet access limits package downloads.

\textbf{Note (v1.0.0):} In-browser firmware flashing via \texttt{esptool-js} works
correctly in the normal use case: the \texttt{esptool-js} library is loaded from its
CDN when the browser first opens the page with an internet connection, and the flash
function then operates fully locally over WebSerial even if internet is subsequently
unavailable. The limitation arises only when the browser is opened on a computer that
has never loaded the page before and has no internet connection at that moment, in
which case the CDN library cannot be fetched. In that scenario, flashing
\texttt{firmware.ino} via the Arduino IDE~\cite{arduinoide} is the fallback; all
remaining pipeline stages (image collection, browser training, weight export, SD
management) continue to work without internet once the page has been loaded. The
portable folder in the repository includes all CDN libraries vendored locally,
allowing the page except firmware flash to operate offline once downloaded.
Use the Arduino IDE offline to load the \texttt{firmware.ino}.

Flashing is handled by \texttt{esptool-js}~\cite{esptooljsrepo}, a WebAssembly port
of Espressif's official flashing tool. Two modes are supported:

\begin{itemize}
  \item \textbf{Option~A --- Sketch only} (\texttt{.ino.bin} at \texttt{0x10000}):
        flashes only the application partition. Fast; preserves the bootloader and NVS
        storage. The correct choice for standard re-flashing once the bootloader is
        present.
  \item \textbf{Option~B --- Full merged binary} (\texttt{.ino.merged.bin} at
        \texttt{0x0}): flashes bootloader, partition table, and application in a single
        pass. Required for fresh chips or after partition changes.
\end{itemize}

\begin{figure}[htbp]
  \centering
  \includegraphics[width=0.50\textwidth]{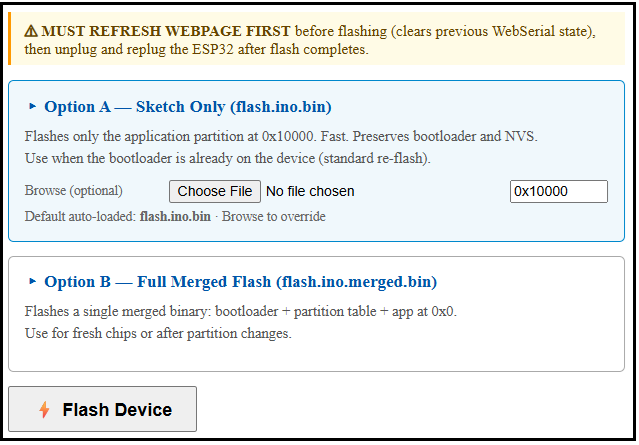}
  \caption{Section~1: In-browser firmware flashing via \texttt{esptool-js}. No Python
           or Arduino IDE is required. Option~A (sketch-only at \texttt{0x10000}) is
           the default for standard re-flashing; Option~B flashes a complete merged
           binary for fresh devices. A progress bar provides real-time feedback;
           the WebSerial monitor reconnects automatically after flashing.}
  \label{fig:flash}
\end{figure}

\subsection{WebSerial Monitor and SD Card Browser}
\label{sec:serial}

Section~2 provides a two-panel interface: a live serial monitor and a graphical SD
card browser, connected to the ESP32 at 115200~baud via the WebSerial API.

The serial monitor displays all firmware output in a scrollable monospace terminal
with optional auto-scroll. Quick-send buttons map to the firmware's single-character
command interface; a free-text command field accepts any protocol string.

The SD browser renders the full directory tree of the ESP32 SD card. Features include
breadcrumb navigation, JPEG inline preview (without downloading to host), inline JSON
and text editing with save-back to the device, and file delete---all without ejecting
the SD card.

\subsubsection{Live \texttt{config.json} Sync}
\label{sec:config}

A \textit{Refresh config.json from SD} button reads \texttt{/header/config.json} and
populates all corresponding browser fields: class labels, learning rate, batch size,
target epochs, augmentation flag, grayscale mode, PSRAM preload flag, and validation
image count. This enables zero-recompile hyperparameter adjustment: the practitioner
edits \texttt{config.json} in the SD browser, saves it back, and the firmware reads
the new values at the next boot without any code change. The schema is shared exactly
between browser and firmware (Listing~\ref{lst:config}).

\begin{lstlisting}[caption={\texttt{config.json} schema shared between browser and
                   firmware. Reading from SD populates all browser training fields;
                   saving back takes effect at the next device boot with no recompile.
                   The \texttt{classLabels} array drives both the ESP32 menu and the
                   browser capture rows, making it the single source of truth for
                   class naming.},
                   label={lst:config}, language={}]
{
  "version": 68,
  "inputSize": 64,
  "numClasses": 3,
  "classLabels": ["0Blank", "1Cup", "2Pen"],
  "learningRate": 0.0003,
  "batchSize": 6,
  "targetEpochs": 20,
  "useAugmentation": true,
  "useGrayscale": false,
  "imagesToPsram": true,
  "validationImages": 3,
  "weightsFile": "myWeights.bin"
}
\end{lstlisting}

\begin{figure}[htbp]
  \centering
  \includegraphics[width=0.50\textwidth]{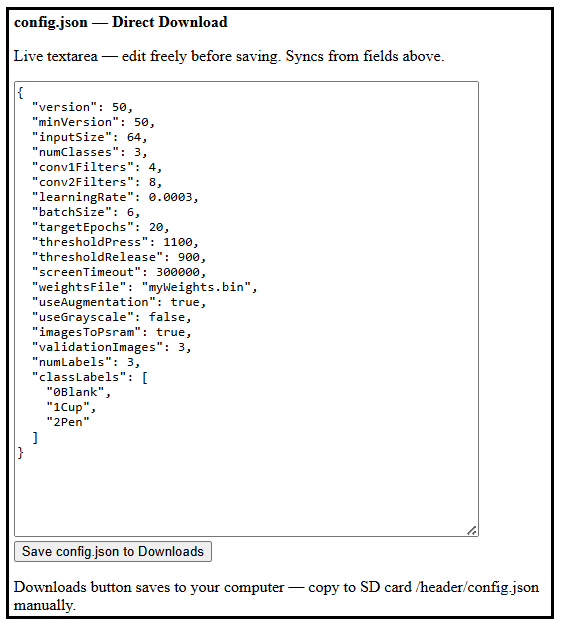}
  \caption{The \texttt{config.json} textarea in Section~5, kept live in sync with
           all training fields. The \texttt{config.json} file can be edited in the
           SD card browser (Section~2) and saved back to the device, or downloaded
           to the PC project folder and copied to the microSD card manually.
           The same JSON is read by the ESP32 firmware at boot, making it the
           single shared configuration source between the browser and the device.}
  \label{fig:sdbrowser}
\end{figure}

\subsection{Camera and Data Collection}
\label{sec:camera}

Section~3 is the data collection pipeline, and its design reflects the TinyML dataset
philosophy described in Section~1.1: images should be collected under the actual
conditions of deployment, not adapted from a general dataset. The practitioner
positions the camera in front of the real objects, under the real lighting, and
captures images that the model will actually encounter at inference time.

Two camera sources are supported:

\begin{itemize}
  \item \textbf{Webcam}: any browser-accessible camera via \texttt{getUserMedia()},
        with device selection from a populated dropdown.
  \item \textbf{ESP32 camera (WebSerial)}: live JPEG frames from the on-device OV2640
        via the \texttt{CAM\_CAPTURE} serial command, arriving as base64-encoded chunks
        and rendered in the browser at a configurable interval.
\end{itemize}

An optional grayscale mode and a resolution selector (64--240~px, with FPS estimates)
must be configured before starting the training pipeline. A live hint displays the
Conv2 output size and flattened dimension for the selected resolution, making the
relationship between input size and model capacity explicit.

\begin{figure}[htbp]
  \centering
  \includegraphics[width=0.50\textwidth]{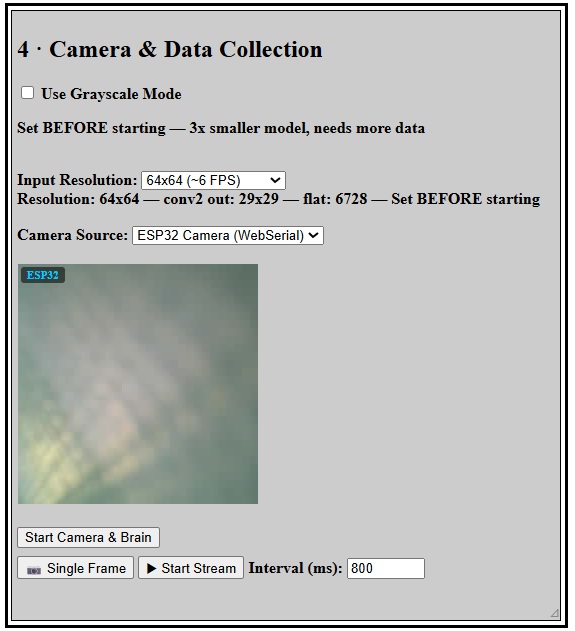}
  \caption{Section~3: Camera data collection. The practitioner captures images of
           their actual objects under actual deployment lighting conditions. For a
           fast iteration cycle, $\sim$30 images per class is sufficient to complete
           training in approximately 1~minute. Images are saved automatically to a
           structured PC project folder; the entire folder is copied to the microSD
           card as a single deployment step. The live preview and per-class sample
           counters give immediate feedback on dataset balance.}
  \label{fig:camera}
\end{figure}

\FloatBarrier
\begin{figure}[htbp]
  \centering
  \includegraphics[width=0.50\textwidth]{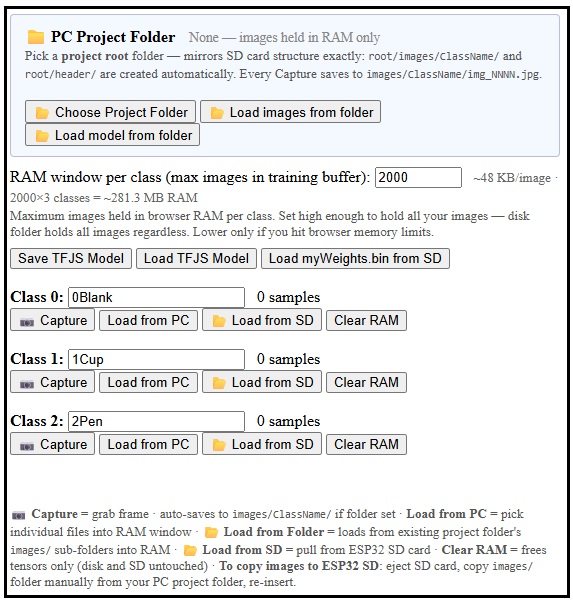}
  \caption{Example training images from the three-class reference dataset (0Blank,
           1Cup, 2Pen) at 64$\times$64 resolution. Images are captured under consistent
           indoor lighting with a plain background. The deliberate simplicity of this
           dataset illustrates the TinyML principle: a clean, task-specific dataset
           outperforms a large general one for a fixed-condition deployment problem.}
  \label{fig:inputs}
\end{figure}

\subsection{Browser-Side CNN Training}
\label{sec:training}

Section~4 runs a CNN in the browser using TensorFlow.js~\cite{tensorflowjs},
GPU-accelerated via WebGL. The architecture exactly mirrors the on-device network
(Table~\ref{tab:arch}).

\begin{table}[h]
\centering
\caption{CNN architecture at \texttt{INPUT\_SIZE}~=~64, identical to the on-device
         firmware. Total parameters: 20,595. The small parameter count is a feature,
         not a limitation: it matches the memory and inference-speed constraints of
         the ESP32-S3 and is sufficient for clean, task-specific classification.}
\label{tab:arch}
\begin{tabular}{@{}lllr@{}}
\toprule
\textbf{Layer} & \textbf{Output shape} & \textbf{Notes} & \textbf{Parameters} \\
\midrule
Input                   & 64$\times$64$\times$3 & RGB, normalized [0,1]     & ---    \\
Conv1 (3$\times$3, 4f)  & 62$\times$62$\times$4 & Leaky ReLU ($\alpha$=0.1) & 112    \\
MaxPool 2$\times$2      & 31$\times$31$\times$4 & ---                       & ---    \\
Conv2 (3$\times$3, 8f)  & 29$\times$29$\times$8 & Leaky ReLU ($\alpha$=0.1) & 296    \\
Flatten                 & 6,728                 & ---                       & ---    \\
Dense (3 classes)       & 3                     & Softmax                   & 20,187 \\
\midrule
\textbf{Total}          &                       &                           & 20,595 \\
\bottomrule
\end{tabular}
\end{table}

The training loop runs on a 150~ms \texttt{setInterval}, interleaved with live
inference so the display remains responsive. Each step assembles a mini-batch (default
size~6), optionally applies stochastic augmentation (brightness jitter and contrast
scaling matching the firmware's augmentation pass), calls
\texttt{model.trainOnBatch()}, and accumulates loss. Every 10 batches the batch
count, average loss, and a qualitative status label (\textit{Improving},
\textit{Converging}, \textit{Well Trained}) are updated.

Two epoch modes are selectable at runtime: \textit{Use All Data} (systematic epoch
ordering) and \textit{random batch mode} (uniform sampling for exploratory training).
Pause and Resume buttons interrupt and restart training without discarding weights or
data.

An \textit{Update Confusion Matrix} button runs the full dataset through the current
model and renders a colour-coded table (diagonal green, off-diagonal red), providing
an immediate per-class accuracy breakdown during or after training.

\FloatBarrier
\begin{figure}[htbp]
  \centering
  \includegraphics[width=0.50\textwidth]{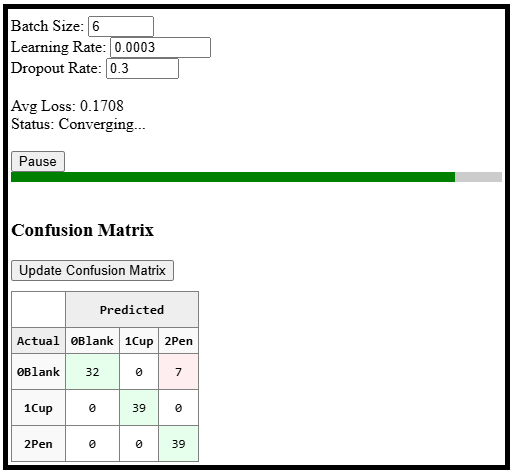}
  \caption{Confusion matrix from a three-class training run (0Blank, 1Cup, 2Pen)
           at 64$\times$64 resolution after 20 epochs. Diagonal cells (green) indicate
           correct classifications; off-diagonal cells (red) identify class confusions.
           The confusion matrix is the primary diagnostic for deciding whether to
           collect more images for a specific class or adjust the training configuration.}
  \label{fig:training}
\end{figure}

\subsection{Weight Export to SD Card}
\label{sec:export}

Section~5 exports browser-trained weights for deployment on the ESP32. The recommended
deployment workflow is: (1) export weights and config to the PC project folder using
the \textit{Export myWeights.bin} and \textit{Save config.json} buttons; (2) copy the
\texttt{header/} subfolder from the PC project folder to the microSD card root; (3)
insert the card and reboot the device. This single folder-copy step deploys both the
trained model and any updated configuration simultaneously. 

\subsubsection{\texttt{myWeights.bin}}

A compact binary file: an ASCII JSON header (architecture metadata and class labels,
bounded by sentinel strings) followed by raw \texttt{float32} arrays in the transposed
layout expected by the ESP32 firmware. The three weight tensors require layout
transposition from TensorFlow.js internal format to the ESP32 C++ loop-order:

\begin{itemize}
  \item Conv1: TF \texttt{[ky, kx, ic, f]} $\rightarrow$ ESP32 \texttt{[f, ky, kx, ic]}
  \item Conv2: TF \texttt{[ky, kx, ic, f]} $\rightarrow$ ESP32 \texttt{[f, ic, ky, kx]}
  \item Dense: TF \texttt{[flat, cls]}     $\rightarrow$ ESP32 \texttt{[cls, f, y, x]}
\end{itemize}

Written to \texttt{/header/myWeights.bin} and loaded automatically at next boot via
the three-tier priority system of Paper~1~\cite{ellis2026ondevice}. The browser can
also load this file from SD at any time, restoring on-device-trained weights into the
live TensorFlow.js model for continued browser-side training or validation.

\subsubsection{\texttt{myWeights.h}}

A self-documenting C header containing all weight arrays as \texttt{float} initialisers
with embedded instructions for enabling \texttt{\#define USE\_BAKED\_WEIGHTS}. This
format is identical to the header the ESP32 firmware itself generates after on-device
training, making the two export paths interchangeable.

\begin{figure}[htbp]
  \centering
  \includegraphics[width=0.50\textwidth]{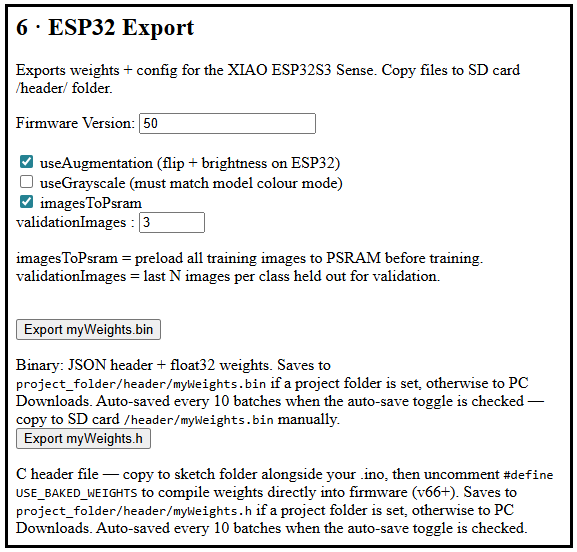}
  \caption{Section~5: Weight and configuration export. \texttt{myWeights.bin} can
           be downloaded to PC for manual transfer to the device SD card. The identical binary format is
           shared with the on-device firmware, making browser-trained weights directly
           deployable at the next device boot.}
  \label{fig:export}
\end{figure}

\section{Conv2 Activation Heatmap}

The live Conv2 activation heatmap is among the most pedagogically distinctive features
of the system. In typical TinyML deployments the internal state of a running model is
entirely opaque: the device produces a class label but exposes no information about
which spatial regions of the input drove that prediction. The heatmap streaming
protocol makes this visible in real-time.

When heatmap mode is enabled, the firmware serialises the maximum activation across
all Conv2 filters at each spatial position into a compact byte array, base64-encodes
it, and transmits it as:

\begin{lstlisting}[caption={Heatmap serial protocol (firmware to browser). One frame
                   is sent per inference cycle while heatmap mode is active.},
                   label={lst:heatmap}, language={}]
HEATMAP:<rows>x<cols>:<base64-bytes>\n
\end{lstlisting}

The browser decodes the payload, maps each byte through a five-segment
blue\,$\rightarrow$\,cyan\,$\rightarrow$\,green\,$\rightarrow$\,yellow\,$\rightarrow$\,red
colour ramp, and renders it on a scaled \texttt{<canvas>} element. At the default
64$\times$64 input, the Conv2 output is 29$\times$29, displayed at 6$\times$ scale
(174$\times$174~px).

The heatmap is also computable entirely in the browser from webcam input without
ESP32 involvement, by extracting Conv2 activations from the TensorFlow.js model
directly. This browser heatmap mode has no impact on ESP32 FPS.

The heatmap serves as a ``glass box'' diagnostic: a practitioner can hold their target
object in front of the camera and observe whether the activation cluster follows the
object or remains anchored to a background artefact---a bright surface, a consistent
light gradient, or an irrelevant texture. If the heatmap shows high activation on the
background rather than the object, the diagnosis is immediate: the dataset lacks
background variation. More images with different backgrounds, or images taken at
different positions, will correct it. To our knowledge, this is one of the first
real-time activation visualizations deployed on a microcontroller-class device and
streamed to a browser for interactive inspection. This feedback loop is unavailable
in other low-cost TinyML deployments we are aware of.

\begin{figure}[htbp]
  \centering
  \includegraphics[width=0.50\textwidth]{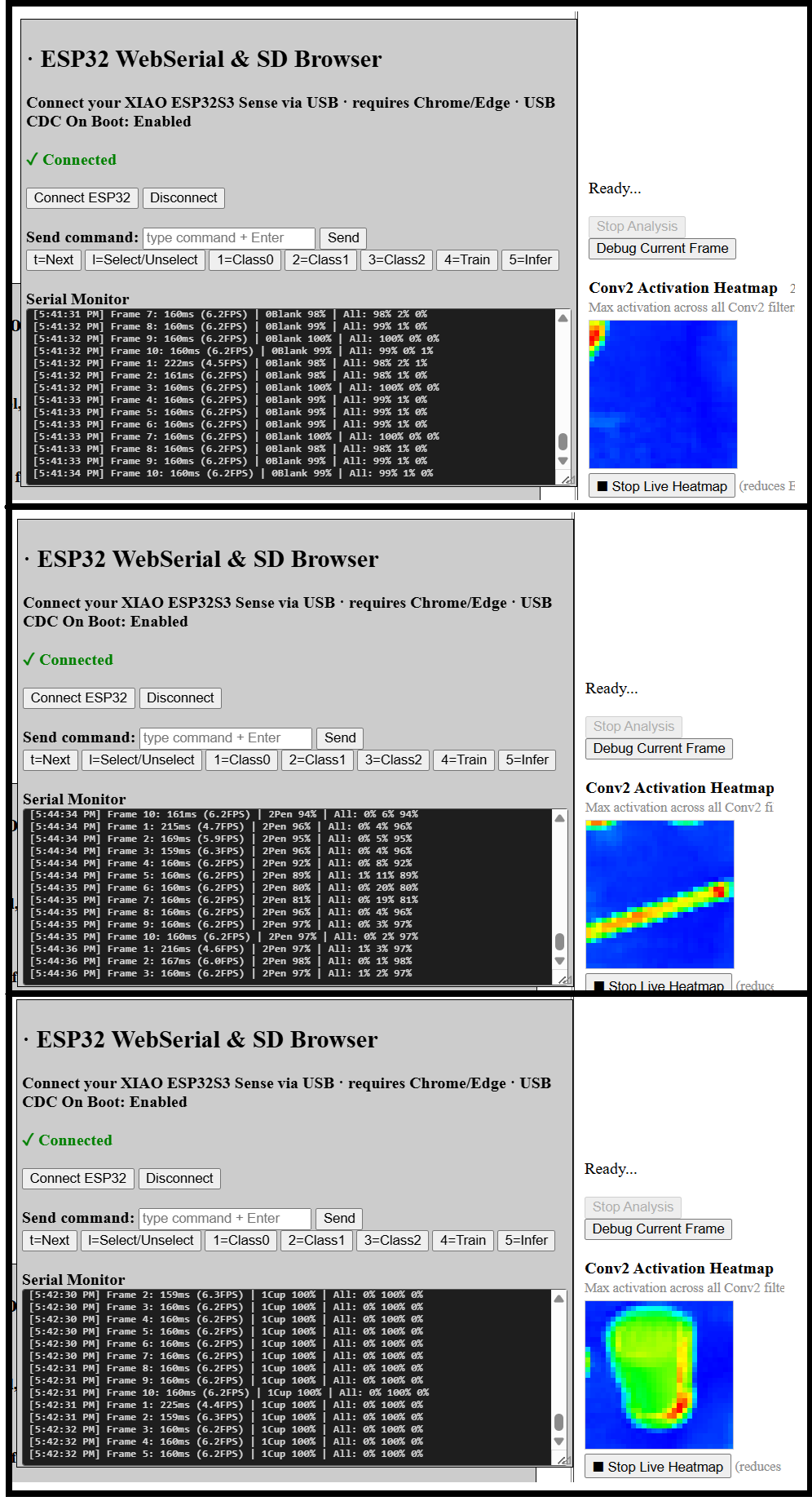}
  \caption{Live Conv2 activation heatmap streamed from the ESP32 during inference.
           Red~=~high activation; blue~=~low activation. The heatmap reveals which
           spatial regions drive the prediction. Here, activation is concentrated on
           the object of interest rather than the background, indicating that the
           model has learned a spatially appropriate feature rather than a spurious
           correlation. If activation instead clusters on the background, the
           practitioner knows immediately that more background-varied images are needed.}
  \label{fig:heatmap}
\end{figure}

\section{Experimental Evaluation}

\subsection{Setup}

All measurements used the Seeed Studio XIAO ESP32-S3 Sense board mounted on the
XIAO ML Kit expansion board (adding the 72$\times$40 OLED display and IMU) with the three-class reference problem (0Blank, 1Cup, 2Pen) at 64$\times$64
resolution under consistent indoor lighting. Browser training ran in Chrome~124 on a laptop with integrated GPU
(WebGL acceleration enabled). On-device results are taken from
Paper~1~\cite{ellis2026ondevice}. Training configuration: batch size~6, learning
rate~0.0003, dropout~0.3, \textbf{100 epochs} per browser run. All raw results,
per-run confusion matrices, real-world inference images, and heatmap captures are
available in the repository at the release tag noted on the title page.

\subsection{Consistency Evaluation: Five Independent Runs}

\begin{figure}[htbp]
  \centering
  \includegraphics[width=0.75\textwidth]{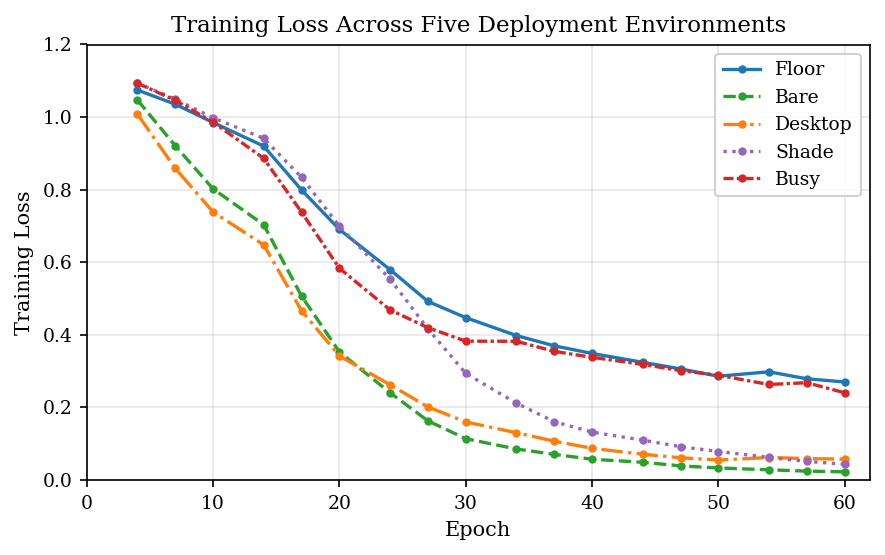}
  \caption{Training loss curves across five independent runs (100 epochs each,
           $\sim$30 images per class, fresh dataset per run). Consistent downward
           convergence across all runs indicates stable training dynamics despite
           dataset variability.}
  \label{fig:training_loss}
\end{figure}

\begin{figure}[htbp]
  \centering
  \includegraphics[width=0.75\textwidth]{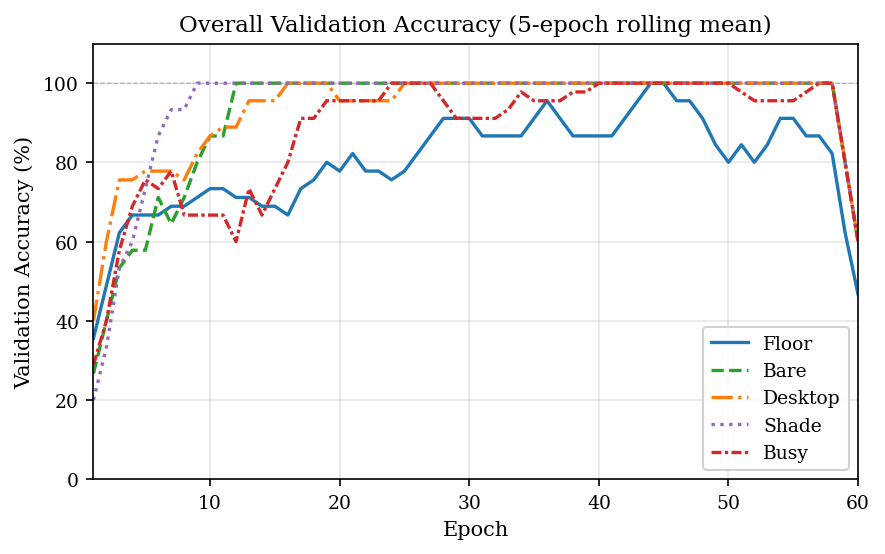}
  \caption{Validation accuracy across five runs. The validation split is held out
           at the start of each run and not used during training. Values are
           reported as percentages for comparability across runs with differing
           image counts.}
  \label{fig:val_accuracy}
\end{figure}

\begin{figure}[htbp]
  \centering
  \includegraphics[width=0.75\textwidth]{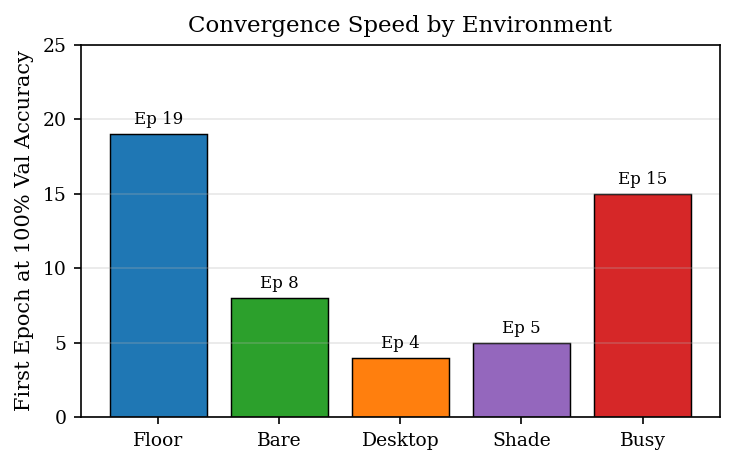}
  \caption{Epoch at which each run first reaches a stable convergence threshold.
           Early convergence in most runs reflects the task-specific nature of the
           dataset: once the model has seen sufficient examples of each class under
           fixed conditions, additional epochs yield diminishing returns.}
  \label{fig:convergence_epoch}
\end{figure}

\begin{figure}[htbp]
  \centering
  \includegraphics[width=0.75\textwidth]{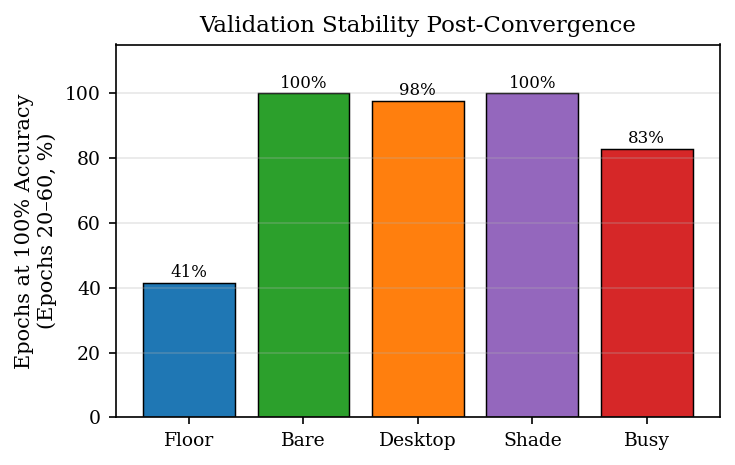}
  \caption{Validation accuracy stability (variance across the last 10 epochs of
           each run) across five test environments: \textit{Floor} (objects placed
           on the floor), \textit{Bare} (plain white paper under good lighting),
           \textit{Desktop} (standard desk surface), \textit{Shade} (very dark
           ambient lighting), and \textit{Busy} (a paint-splattered workbench).
           Low variance indicates that training has reached a stable plateau; high
           variance suggests the model has not fully converged or the dataset is
           too small for the selected epoch count.}
  \label{fig:val_stability}
\end{figure}

To assess training stability across dataset variation, five training runs were
conducted with independently collected image sets (fresh captures per run, $\sim$30
images per class). Results are summarised in Table~\ref{tab:multirun} and
Figures~\ref{fig:training_loss}--\ref{fig:val_stability}.

\textbf{Validation set note (v1.0.0):} In the initial experimental design, the
validation split was fixed at 3 images per class regardless of training set size.
As the training set grew from the original 9~images/class to $\sim$30~images/class
and beyond, the validation set remained at 3~images/class, producing an increasingly
unbalanced train-to-validation ratio. A future version of these experiments will use
a proportional validation split (e.g., 20\% held-out per class) to provide more
statistically stable accuracy estimates as dataset size increases.

\begin{table}[h]
\centering
\caption{Five-run consistency evaluation on the three-class reference problem
         (0Blank, 1Cup, 2Pen). Each run used a freshly collected dataset of
         $\sim$30 images per class, 100 epochs, browser training (TensorFlow.js,
         WebGL). Percentages are reported for cross-run comparability; raw
         counts are available in the repository. Confusion matrices are computed
         on the held-out validation split.}
\label{tab:multirun}
\begin{tabular}{@{}lccc@{}}
\toprule
\textbf{Metric} & \textbf{Mean} & \textbf{Std} & \textbf{Notes} \\
\midrule
Final training loss        & 0.127 & 0.118   & See Fig.~\ref{fig:training_loss} \\
Final validation accuracy  & 95.6\% & 9.9\%  & See Fig.~\ref{fig:val_accuracy} \\
Convergence epoch          & 10.2  & 6.5     & See Fig.~\ref{fig:convergence_epoch} \\
Real-world inference Blank & 87\%  & 14\%   & See Fig.~\ref{fig:val_vs_realworld} \\
Real-world inference cup   & 98.0\% & 2.1\% & See Fig.~\ref{fig:val_vs_realworld} \\
Real-world inference pen   & 83.0\% & 0.7\% & See Fig.~\ref{fig:val_vs_realworld} \\
\bottomrule
\end{tabular}
\end{table}

\subsection{Training Accuracy vs.\ Real-World Inference}
\label{sec:realworld}

A consistent finding across all five runs is that training and validation confusion
matrices---which are computed on images captured under the same conditions as the
training set---overstate real-world inference accuracy. When the model is evaluated
on live camera images during deployment, accuracy can be lower, particularly for
classes whose features are sensitive to small changes in distance, angle, or ambient
light level.

\begin{figure}[htbp]
  \centering
  \includegraphics[width=0.75\textwidth]{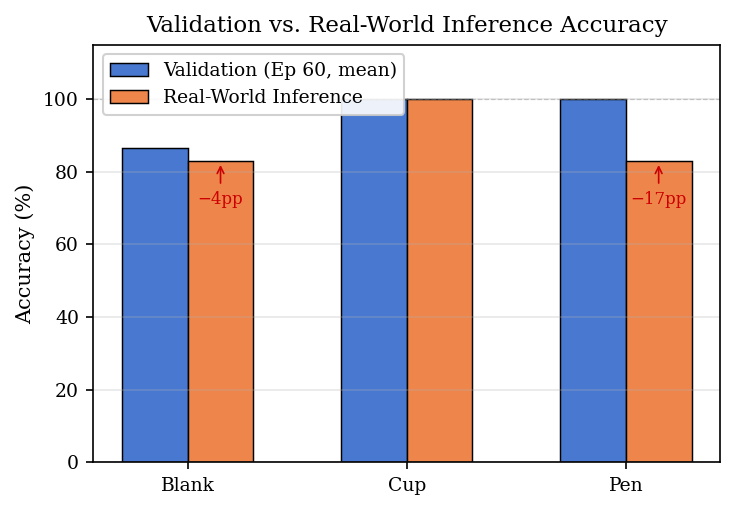}
  \caption{Validation accuracy (computed on held-out training-set images) versus
           real-world inference accuracy (computed on live camera images at deployment
           time) across five runs.
           The gap reflects the practitioner's own deployment conditions, not a flaw
           in the training pipeline. Closing this gap is achieved by capturing training
           images that more faithfully represent the range of positions and lighting
           variations the device will encounter.}
  \label{fig:val_vs_realworld}
\end{figure}

Figure~\ref{fig:val_vs_realworld} summarises this gap across runs.

\begin{figure}[htbp]
  \centering
  \includegraphics[width=0.45\textwidth]{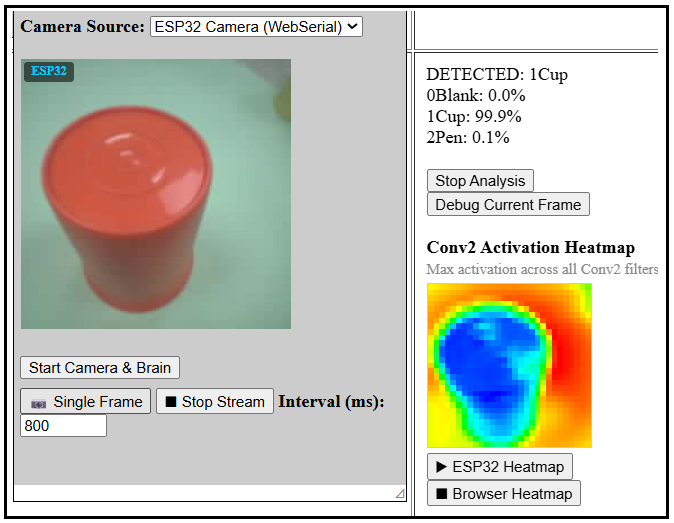}
  \caption{Live inference on the cup class under normal lighting: raw camera
           image (left), Conv2 activation heatmap (centre), and per-class confidence
           scores (right). The model correctly classified the cup with 99.9\%
           confidence. The heatmap confirms that activation is concentrated on
           the cup object rather than the background, indicating that the model
           has learned a spatially appropriate feature. The cup class was the most
           consistently learnable across all five runs, likely because its circular
           rim and solid body produce a distinctive Conv2 response even at small
           image sizes.}
  \label{fig:bare_cup}
\end{figure}

This gap is both expected and informative. The confusion matrix is the correct tool
for diagnosing \emph{which classes} are confused during training; the real-world
inference images (with live heatmaps, see Figure~\ref{fig:bare_cup}) are the correct
tool for diagnosing whether the model has learned the object or a background artefact.
A high confusion-matrix score with low real-world accuracy is a signal to collect
images under more varied conditions---different distances, angles, and lighting levels.

An important secondary finding is that the models can be trained effectively under
low light conditions. Provided the training images are captured under the same
low-light conditions as deployment, the model generalises to that environment reliably.
This is consistent with the deployment philosophy of Section~\ref{sec:deployment_philosophy}:
dataset representativeness of the actual deployment conditions matters more than
absolute image quality.

\begin{figure}[htbp]
  \centering
  \includegraphics[width=0.45\textwidth]{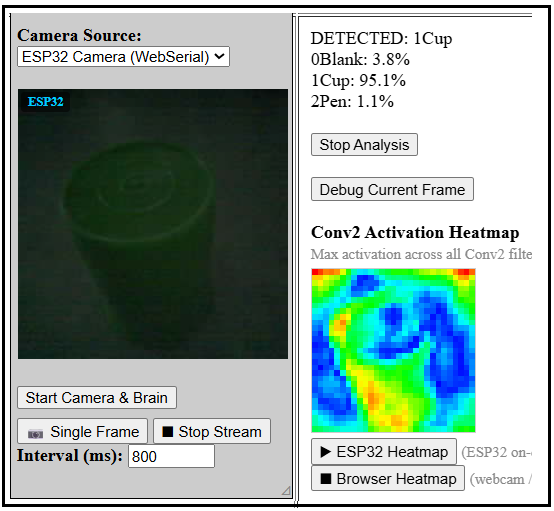}
  \caption{Live inference on the cup class under very low ambient light: raw camera
           image (left), Conv2 activation heatmap (centre), and per-class confidence
           scores (right). Despite the extremely dark image, the model correctly
           classified the cup with 95\% confidence. The heatmap confirms that
           activation remains concentrated on the cup object rather than the
           background, demonstrating that a task-specific model trained under
           matching low-light conditions generalises reliably to that environment.
           This result illustrates the core TinyML deployment philosophy: dataset
           representativeness of the actual deployment conditions---including
           challenging lighting---matters more than absolute image brightness.}
  \label{fig:shade_cup}
\end{figure}

\subsection{Training Speed and Accuracy}

Table~\ref{tab:eval} presents the key performance comparison. Browser-side TensorFlow.js
training reaches the same approximate loss as a full on-device run ($\sim$9 minutes) in
approximately 1~minute for a representative dataset of $\sim$30 images per class---a
$9\times$ reduction in per-run iteration time. A complete collect--train--deploy cycle
(image capture, browser training, weight export to PC folder, microSD copy, device
reboot) is achievable in under 10~minutes, enabling several full iterations within a
single class period. This speed difference stems from WebGL GPU parallelism versus the
single-core floating-point unit of the ESP32-S3.

Browser training also typically reaches slightly higher accuracy than on-device training
on the same dataset, because the WebGL backend provides a smoother optimization
landscape than the ESP32 fixed-point FPU approximation. Weights exported from the
browser and loaded onto the device perform at the browser-trained accuracy.

\begin{table}[h]
\centering
\caption{Performance comparison: on-device (Paper~1) vs.\ browser-side training
         (this work). Reference dataset: 3 classes, $\sim$30 images/class,
         \texttt{INPUT\_SIZE}~=~64, 100 epochs. The $9\times$ training speed improvement
         is the primary new result; the weight format and file size are identical
         between platforms.}
\label{tab:eval}
\begin{tabular}{@{}p{5cm}p{3.5cm}p{3.5cm}l@{}}
\toprule
\textbf{Metric} & \textbf{On-device (Paper~1)} & \textbf{Browser (this work)} & \textbf{Notes} \\
\midrule
Training time to equiv.\ loss  & $\sim$9 min     & $\sim$1 min    & $9\times$ faster \\
Full collect--train--deploy cycle & $\sim$30 min & $<$3 min       & incl.\ SD copy \\
Typical training accuracy       & 71--79\%       & 75--88\%       & Browser has WebGL \\
Validation accuracy (3-class)   & 67--100\%      & 67--100\%      & Same held-out split \\
Inference rate (device)         & 6.3 FPS        & 6.3 FPS        & Runs on ESP32-S3 \\
Heatmap FPS cost (device)       & $\sim$0.2 FPS  & $\sim$0.2 FPS  & Serial overhead \\
Weight file (\texttt{.bin})     & $\sim$83 KB    & $\sim$83 KB    & Identical format \\
Firmware flash time             & $\sim$2 min (Arduino IDE) & $\sim$30 s  & Zero install \\
\bottomrule
\end{tabular}
\end{table}

\subsection{Dataset Size and Sufficiency}
\label{sec:deployment_philosophy}

A recurring question in TinyML education is: how many images are enough? For a
fixed-condition, specific-task classifier on this architecture, the answer is
substantially smaller than general computer vision intuition suggests.
Table~\ref{tab:datasize} shows accuracy across dataset sizes for the three-class
reference problem.

\begin{table}[h]
\centering
\caption{Validation accuracy vs.\ dataset size per class for the 3-class reference
         problem (0Blank, 1Cup, 2Pen), browser training, 64$\times$64 RGB,
         100 epochs, consistent lighting and background.}
\label{tab:datasize}
\begin{tabular}{@{}lll@{}}
\toprule
\textbf{Images/class} & \textbf{Training accuracy} & \textbf{Validation accuracy} \\
\midrule
9  (Dataset~A)         & 65--72\% & 55--67\% \\
$\sim$20               & 70--80\% & 65--75\% \\
38 (Dataset~B)         & 75--88\% & 70--85\% \\
100                    & 80--90\% & 75--88\% \\
\bottomrule
\end{tabular}
\end{table}

The key insight is that these models are \emph{intentionally} trained to overfit to
their deployment conditions, which is desirable in fixed-condition TinyML applications.
A model that achieves 100\% training accuracy on its specific task under its specific
conditions is behaving correctly; the relevant measure is real-world inference accuracy
under those same conditions. Dataset \emph{quality} and \emph{representativeness of
deployment conditions} matter more than quantity. A dataset of 60 images collected
under the exact lighting and camera position of the deployment outperforms a dataset
of 200 images collected under varied conditions that do not match deployment. This is
the defining characteristic of TinyML data collection, and it separates TinyML from
edge AI more broadly: edge AI seeks to bring general AI capability to local hardware;
TinyML seeks to perform a specific task at minimum energy and cost. Both are valid
goals, and both are served by this system.

\section{WebSerial Communication Protocol}

Table~\ref{tab:protocol} summarises the full set of WebSerial commands exchanged
between the browser and the firmware. All commands are newline-terminated ASCII strings
sent at 115200~baud. The protocol uses explicit \texttt{\_START}/\texttt{\_END}
sentinels for all multi-line responses so the browser parser can reassemble responses
without relying on timing. An incomplete transaction (e.g., from a device reset) is
detected by the absence of the closing sentinel and discarded.

\begin{table}[h]
\centering
\caption{WebSerial command protocol between browser and ESP32 firmware.}
\label{tab:protocol}
\begin{tabular}{@{}p{4.3cm}p{4.0cm}p{5.7cm}@{}}
\toprule
\textbf{Browser $\rightarrow$ ESP32} & \textbf{ESP32 $\rightarrow$ Browser} & \textbf{Purpose} \\
\midrule
\texttt{CAM\_CAPTURE:WxH:Q}
  & \texttt{CAM\_JPEG\_START} $\to$ \texttt{CAM\_JPEG:}\ldots $\to$ \texttt{CAM\_JPEG\_END}
  & Request one JPEG frame \\[4pt]
\texttt{CAM\_STREAM\_STOP}
  & ---
  & Stop continuous stream \\[4pt]
\texttt{SD\_LIST:/path}
  & \texttt{SD\_LIST\_START} $\to$ \texttt{SD\_FILE:}\ldots $\to$ \texttt{SD\_LIST\_END}
  & Directory listing \\[4pt]
\texttt{SD\_READ:/path}
  & \texttt{SD\_CONTENT\_START} $\to$ \texttt{SD\_LINE:}\ldots $\to$ \texttt{SD\_CONTENT\_END}
  & Read text file \\[4pt]
\texttt{SD\_JPEG:/path}
  & \texttt{SD\_JPEG\_START} $\to$ \texttt{SD\_JPEG:}\ldots $\to$ \texttt{SD\_JPEG\_END}
  & Read JPEG from SD \\[4pt]
\texttt{SD\_JPEG\_WRITE\_START:/p:N} \newline
\texttt{SD\_JPEG\_CHUNK:b64} $\times N$ \newline
\texttt{SD\_JPEG\_WRITE\_END}
  & \texttt{OK:JPEG\_WRITE\_DONE /path (NB)}
  & Write any binary or text file \\[4pt]
\texttt{SD\_DELETE:/path}
  & \texttt{OK:Deleted}
  & Delete file \\[4pt]
\texttt{SD\_RMDIR:/path}
  & \texttt{OK:Deleted}
  & Delete directory (recursive) \\[4pt]
\texttt{HEATMAP\_ON}
  & \texttt{HEATMAP:R}$\times$\texttt{C:b64} per frame
  & Enable Conv2 heatmap stream \\[4pt]
\texttt{HEATMAP\_OFF}
  & ---
  & Disable heatmap stream \\[4pt]
\texttt{1}--\texttt{5}, \texttt{t}, \texttt{l}
  & (firmware menu state change)
  & Direct firmware menu navigation \\
\bottomrule
\end{tabular}
\end{table}

\section{Comparison to Paper~1 Workflow}

Table~\ref{tab:comparison} summarises the capabilities added by the browser companion
relative to the standalone on-device workflow of Paper~1.

\begin{table}[h]
\centering
\caption{Capability comparison: on-device (Paper~1) vs.\ WebSerial companion (Paper~2).}
\label{tab:comparison}
\begin{tabular}{@{}p{4.2cm}p{4cm}p{4.5cm}@{}}
\toprule
\textbf{Capability} & \textbf{Paper~1 (on-device)} & \textbf{Paper~2 (browser)} \\
\midrule
Firmware installation   & Arduino IDE required          & Browser flash via \texttt{esptool-js} (Arduino IDE fallback for cold-offline start) \\
Data collection         & On-device touch / SD          & Webcam or ESP32 camera in browser \\
Training time           & $\sim$9 min/run (C++ FPU)     & $\sim$1 min/run (TF.js + WebGL); full cycle $<$10~min \\
SD file management      & SD card ejection required     & Full browser SD browser \\
Config adjustment       & Recompile required            & Live sync, zero recompile \\
Weight export           & SD card only                  & SD direct or PC download \\
Weight import           & SD boot auto-load             & SD or PC upload \\
Confusion matrix        & Serial text output            & Colour-coded HTML table \\
Conv2 heatmap           & Not available                 & Live streaming from firmware \\
Energy measurement      & External PPK2 profiler        & Not applicable (laptop training) \\
Raw data privacy        & Never leaves device           & Stays on local machine \\
LLM adaptability        & Full C++ source               & Single HTML file \\
\bottomrule
\end{tabular}
\end{table}

The two approaches are complementary. Paper~1 is the correct choice when the
microcontroller must operate untethered, when raw data must never leave the device, or
when understanding the C++ training loop at the instruction level is the pedagogical
goal. Paper~2 is the correct choice when faster iteration, a richer management
interface, or a lower barrier to first-session entry is the priority.

\section{Limitations}

\begin{itemize}
  \item \textbf{Browser compatibility.} WebSerial is supported only in Chromium-based
        browsers (Chrome~89+, Edge~89+). Firefox and Safari are not supported.
        The page runs fully locally by opening \texttt{index.html} directly in
        Chrome or Edge --- no local HTTP server is required. The portable folder
        in the repository vendors all CDN libraries locally, so the complete
        page (excluding firmware flash) operates offline once that folder has been
        downloaded. WebSerial itself always requires a Chromium engine regardless
        of how the page is served.

  \item \textbf{SD card deployment via PC folder copy (recommended); browser-side SD write not yet active.}
        The primary recommended workflow is to save all trained artefacts---images, weights
        (\texttt{myWeights.bin}, \texttt{myWeights.h}), and \texttt{config.json}---to
        a PC project folder whose directory layout mirrors the SD card root exactly.
        The entire folder is then copied to the microSD card in a single manual operation
        before inserting into the device. This avoids serial throughput constraints
        entirely. The \textit{Export myWeights.bin} button writes directly to the
        PC folder. A planned browser feature would automate this PC-to-SD transfer
        without manual card handling, but this capability is not active in v1.0.0.

  \item \textbf{No persistent in-browser model storage.} TensorFlow.js weights are
        held in browser memory and lost on page refresh unless explicitly exported via
        the export buttons. The \textit{Save TFJS Model} button provides a JSON +
        binary download for browser-portable saves; \textit{Export myWeights.bin} is
        the recommended path for device deployment.

  \item \textbf{Architecture lock-in.} The CNN filter counts (Conv1~=~4, Conv2~=~8)
        are fixed to match the on-device firmware. Changing filter counts requires
        coordinated modification of both the JavaScript export transpositions and the
        firmware \texttt{\#define} constants. This is explicitly a supported use case
        for LLM-assisted modification: the transposition logic and relevant constants
        are documented here and in the source file.

  \item \textbf{Model capacity.} The 20,595-parameter architecture is sufficient for
        simple classification on clean, task-specific datasets but is not suitable for
        complex or background-varied scenes. For harder problems, the recommended path
        is better data---more varied images under deployment conditions---rather than a
        larger model, because model size is bounded by the ESP32's available PSRAM for
        on-device inference.

  \item \textbf{CDN dependency.} The three JavaScript libraries (TensorFlow.js, JSZip,
        esptool-js) are loaded from public CDNs when the page is first opened with
        internet access. Once loaded, they are cached by the browser and the page
        operates fully offline. For environments where the browser cannot be pre-loaded
        with internet, the portable folder in the repository includes all three
        libraries vendored locally alongside \texttt{index.html}, requiring no CDN
        access at any point, except for firmware flash.
\end{itemize}

\section{Repository and Reproducibility}

The application is maintained at:
\begin{center}
\url{https://github.com/webmcu-ai/webmcu-vision-web}
\end{center}

This paper corresponds to release tag \texttt{v1.0.0}:
\begin{center}
\url{https://github.com/webmcu-ai/webmcu-vision-web/releases/tag/v1.0.0}
\end{center}

The release includes \texttt{index.html}, firmware.ino, companion firmware binaries
(\texttt{.ino.bin}, \texttt{.ino.merged.bin}), all five-run evaluation datasets,
per-run confusion matrices, real-world inference images with heatmaps, portable folder structure, and the
figure source data. No build step is required: open \texttt{index.html} in Chrome
or Edge and connect the ESP32 via USB.

The on-device firmware (Paper~1) is at:
\begin{center}
\url{https://github.com/webmcu-ai/on-device-vision-ai}
\end{center}

\textbf{Reproducibility checklist:}
\begin{itemize}
  \item Hardware: Seeed Studio XIAO ESP32-S3 Sense (8~MB PSRAM variant) with XIAO ML Kit
        expansion board (MicroSD slot, OLED display).
  \item Firmware: \texttt{webmcu-vision-web} v71 (\texttt{firmware.ino}), compiled
        with Arduino IDE 2.x, board package \textit{esp32} v3.x, \texttt{U8g2} v2.x.
  \item Browser application: \texttt{webmcu-vision-web} \texttt{index.html} (release \texttt{v1.0.0});
        TensorFlow.js 4.22.0, esptool-js 0.5.7, JSZip 3.10.1.
  \item Browser: Chrome~124+ or Edge~124+.
  \item Reference dataset: $\sim$30 images/class per run; 3 classes (0Blank, 1Cup, 2Pen);
        5 independent runs with freshly collected images; consistent indoor lighting.
  \item Training configuration: \texttt{INPUT\_SIZE}~=~64, batch~6,
        lr~0.0003, 100 epochs, dropout~0.3.
  \item Deployment: save artefacts to PC project folder via \textit{Export myWeights.bin}
        and \textit{Save config.json}; copy \texttt{header/} folder to microSD card;
        reinsert and reboot device.
  \item All evaluation artefacts (per-run datasets, confusion matrices, inference images,
        heatmaps, figure source data) available at the \texttt{v1.0.0} release tag.
\end{itemize}

Both repositories are released under the MIT License.

\section{Future Work}

Paper~3 of this series will extend the framework to on-device audio classification
using the XIAO ML Kit's built-in I2S microphone, with a companion browser application
providing spectrogram visualization and audio data collection. Paper~4 will extend to
IMU-based gesture and activity recognition using the ML Kit's on-board IMU. Both will
follow the same template structure and be documented for LLM-assisted adaptation.

Planned additions to \texttt{webmcu-vision-web} include:

\begin{itemize}
  \item \textbf{Proportional validation split.} Future experimental runs will use a
        fixed percentage (e.g., 20\%) of each class as the held-out validation set,
        rather than the fixed 3-image split used in v1.0.0. This ensures that
        validation accuracy estimates remain statistically meaningful as training
        datasets scale beyond $\sim$30 images per class.

  \item \textbf{Offline-first firmware flashing.} In-browser firmware flashing via
        \texttt{esptool-js} works correctly when the page has been loaded with internet
        access. The portable folder already addresses the CDN dependency for offline use.
        A future improvement will ensure the flash workflow is clearly documented and
        tested for first-boot offline scenarios, and will explore caching strategies
        so that the flash library is retained across browser sessions without relying
        on the portable folder.

  \item \textbf{Manual PC-to-SD card transfer.} The current workflow requires
        manually copying the PC project folder to the microSD card. A planned browser
        feature will automate this transfer step, eliminating the need to eject and
        reinsert the card between training and deployment, but this capability is not
        active in v1.0.0.
  \item Dynamic multi-class UI generation driven by \texttt{numClasses} read from
        \texttt{config.json}, removing the current three-class hard-coding.
  \item INT8 quantization export for compatibility with Espressif's ESP-NN production
        inference kernels~\cite{espressifespdl}, providing a natural follow-on exercise
        for students who have understood the full-precision baseline.
  \item WebGPU acceleration (replacing WebGL) for further training speed improvements
        as WebGPU browser support matures.
  \item Offline-capable bundle (all CDN libraries vendored locally) for use in
        restricted-network classrooms and field deployments in the global south.
  \item Percentage/count toggle in the confusion matrix display, aligning the in-browser
        diagnostic view with the percentage-normalized values used in published results.
\end{itemize}

\paragraph{LLM-adaptable design.}
The single-file architecture has a secondary benefit beyond zero-install deployment:
it fits within the context window of current LLM assistants, making the entire
codebase available for targeted modification in a single conversation. A practitioner
can describe a new sensor modality, hardware variant, or classification task to an
LLM and receive a targeted diff to the relevant section of the file. The fixed naming
conventions, consistent comment structure, and self-contained module layout are all
chosen to make these modifications deterministic and verifiable. This design principle
extends to the paper series itself: each paper documents its companion repository
precisely enough that an LLM-assisted practitioner can reproduce, adapt, and extend
the system without deep software engineering expertise.

Community contributions are welcomed as GitHub issues and pull requests at
\url{https://github.com/webmcu-ai} under the MIT License.

\section{Conclusion}

\texttt{webmcu-vision-web} is a complete, private, browser-based machine learning
training and deployment system for TinyML vision on the Seeed Studio XIAO ESP32-S3
Sense as the XIAO ML kit. In a single HTML file it provides: zero-install firmware flashing; a live
serial monitor; a full SD card browser with image preview and inline editing;
zero-recompile hyperparameter adjustment via \texttt{config.json} live-sync;
webcam and ESP32 OV2640 camera data collection for task-specific datasets;
TensorFlow.js CNN training completing in approximately 1~minute for a representative
three-class dataset of $\sim$30 images per class; a confusion matrix; weight export
in binary and C header formats; and live Conv2 activation heatmap streaming from the
running device.

Together with the concepts from the on-device Arduino firmware of Paper~1, the system covers the full
embedded ML lifecycle: flash the device from the browser, collect images under your
exact deployment conditions, train in the browser, save all artefacts to a PC project
folder, copy the folder to the microSD card, and reboot the device to run inference
at 6.3~FPS. This complete collect--train--deploy cycle takes under 10~minutes for a
three-class problem, and can be repeated multiple times within a single class period
or work session.

A central design principle is that TinyML works best when the dataset matches the
deployment conditions exactly. These models are intentionally trained to overfit to
their specific deployment environment---a specific object, under specific lighting,
at a specific position. This is a feature of TinyML, not a limitation: real-world
inference accuracy is the relevant measure, and it is maximised by capturing training
images that faithfully represent the deployment context. The five-run consistency
evaluation presented here demonstrates this principle in practice and provides
reproducible baselines at the v1.0.0 release tag.

The repository is structured as a living template for LLM-assisted adaptation,
enabling educators, small businesses, and researchers in any environment to modify the
system for their specific hardware, sensors, and classification tasks. No data leaves
the local machine at any stage. All source is MIT-licensed at
\url{https://github.com/webmcu-ai/webmcu-vision-web}.

\section*{Acknowledgements}

The author thanks Brian Plancher (Dartmouth College) for his generous support and
suggestions during this project; his contributions to the TinyML4D community~\cite{plancher2024tinymld}
and to Paper~1 of this series provided both technical grounding and useful context for the
browser-based companion described here.

The author used free tier LLM assistants---Claude (Anthropic), ChatGPT (OpenAI), and Gemini
(Google)---for structural editing, code review and research, and section-level refactoring of the
single-file application; all technical decisions, experimental results, and
architectural choices are the author's own. The LLM-assisted workflow described in
this paper is itself a demonstration of the approach: a practitioner with domain
knowledge of the hardware and the problem can use LLM tools to accelerate development
of a working embedded ML system without requiring deep software engineering expertise.

\end{document}